\pdfoutput=1

\documentclass[11pt]{article}

\usepackage{acl}

\usepackage{fancyhdr}
\usepackage{times}
\usepackage{latexsym}
\usepackage[T1]{fontenc}

\usepackage[utf8]{inputenc}

\usepackage{microtype}
\usepackage{graphicx}

\title{FabKG: A Knowledge graph of Manufacturing Science domain utilizing structured and unconventional unstructured knowledge source }

\author{Aman Kumar \\
  akumar33@ncsu.edu \And
  Akshay G Bharadwaj \\
  abharad3@ncsu.edu \And
  Binil Starly \\
  bstarly@ncsu.edu \And
  Collin Lynch \\
  cflynch@ncsu.edu 
  }

\begin{document}
\maketitle
\begin{abstract}
As the demands for large-scale information processing have grown, knowledge graph-based approaches have gained prominence for representing general and domain knowledge. The development of such general representations is essential, particularly in domains such as manufacturing which intelligent processes and adaptive education can enhance. Despite the continuous accumulation of text in these domains, the lack of structured data has created information extraction and knowledge transfer barriers. In this paper, we report on work towards developing robust knowledge graphs based upon entity and relation data for both commercial and educational uses. To create the FabKG (Manufacturing knowledge graph), we have utilized textbook index words, research paper keywords, FabNER (manufacturing NER), to extract a sub knowledge base contained within Wikidata. Moreover, we propose a novel crowdsourcing method for KG creation by leveraging student notes, which contain invaluable information but are not captured as meaningful information, excluding their use in personal preparation for learning and written exams. We have created a knowledge graph containing 65000+ triples using all data sources. We have also shown the use case of domain-specific question answering and expression/formula-based question answering for educational purposes.
\end{abstract}

\section{Introduction}

In recent years, the advancement of artificial intelligence applications has grown multifold. Many areas such as natural language processing, digital twins \cite{liu2021multi}, and chatbots \cite{chen2021multi} have become very popular for their ability to record and use information from unstructured sources efficiently. One such application is Knowledge Graph (KG), which has gained popularity in various domains due to its potential applications. A Knowledge Graph is a data graph meant to accumulate and impart real-world knowledge, with nodes representing entities of interest and edges representing potentially diverse relations between the entities. A KG has varied applications in recommendations, search, question answering and many more. Most importantly, a KG can be used to make decisions based on inferences.  

The use of a knowledge graph is of high value in making design and manufacturing-related decisions. As there has been an explosion of knowledge addition in various design considerations and manufacturing decisions, most of the knowledge is with Small and medium-sized enterprises (SMEs). The decision-making in design and production could be significantly improved using knowledge graphs \cite{buchgeher2021knowledge}. It can benefit not only small and medium manufacturers \cite{li2021design}, but also hardware-based entrepreneurs and help boost self-sustaining product development \cite{li2020knowledge}.  

\begin{figure*}
    \centering
    \includegraphics[width=\textwidth,height=5cm]{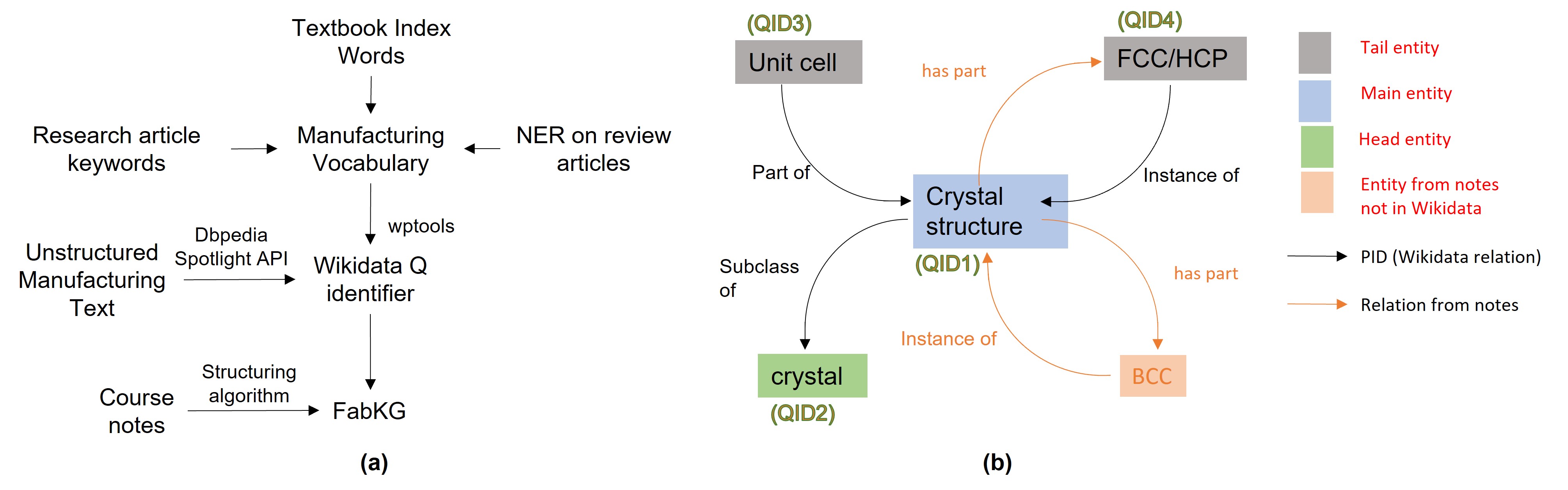}
    \caption{(a) Manufacturing Knowledge Graph construction methodology (b) Use of SPARQLWrapper to fetch wikidata items associated with 'crystal structure' in two step forward and two step backward. This image also shows addition of entities from student notes. }
    \label{figure1}
\end{figure*}

A number of prior researchers have started developing manufacturing related knowledge graphs based on specific problem areas such as machining process planning \cite{yang2019knowledge, ye2018design}, workshop resource KG \cite{zhou2021novel, sun2019multi}, intelligent manufacturing \cite{yan2020knowime},  faults  \cite{liang2022pf2rm, wang2019tutorial}, maintenance \cite{hossayni2020semkore} and industry 4.0 \cite{garofalo2018leveraging, bader2020knowledge, kraft2020j}.  However, none of these graphs represent fundamental knowledge of manufacturing concepts, processes, process parameters, characterization, materials, applications, and various other basic aspects of manufacturing domain education. A large amount of such fragmented knowledge can be integrated to assist the learners in intuitively and easily connecting with the knowledge system by leveraging the nodes and relationships. Such knowledge integration will also assist in intelligent question answering that can accelerate knowledge discovery and search.

Google bases part of its Knowledge Vault on the well-known Wikidata knowledge base \cite{ringler2017one}. Even though Wikidata has a large amount of information from Wikipedia, there is a dearth of standardized knowledge regarding many important entities related to the Manufacturing domain. For instance, the term ‘additive manufacturing’ is present as ‘3d printing’; while there have been substantial developments in the field of ‘metal additive manufacturing’ (metal AM) over the last decade, it is not present as a subclass of ‘3d printing’ in Wikidata. Moreover, within metal additive manufacturing \cite{frazier2014metal}, sub-classifications such as DMLS, EBAM, and PBF are not present in Wikidata. One reason for this is the volunteer-driven nature of Wikidata as a knowledge base; this has led to a limited amount of specialist terminology and information regarding the manufacturing domain. Therefore, Wikidata cannot provide direct answers to questions that are very specific to this domain. To understand the basic concepts in the context of manufacturing we focus on formulating answers to some basic questions such as, ‘What are some precision finishing manufacturing process?’, ‘What are some tools for machining copper?’ etc. The purpose of creating such a knowledge graph of manufacturing using Wikidata is to provide a starting point for a structured manufacturing knowledge base, which can be amalgamated with knowledge from other sources such as textbook \cite{rahdari2020using} and research articles \cite{wang2020covid}. 

To tackle the challenges in developing the knowledge graph from scratch for manufacturing science, we consider various methodologies for creating accurate graphs. We propose a merged knowledge graph that combines the existing structured Wikidata knowledge graph with a novel semi-supervised knowledge graph extracted from textbook data. For extracting graph triples from Wikidata, as mentioned in Figure \ref{figure1}, we have adopted two methods for the approach: (1) Vocabulary-based and (2) Based on Unstructured text. Former includes fetching Wikidata items using a collection of manufacturing vocabulary terms through the utilization of textbook index words, keywords from research papers, and named entity recognition using FabNER \cite{kumar2021fabner}, followed by the use of DBpedia \cite{mendes2011dbpedia} to find Wikidata items. The latter is a semi-supervised approach that utilizes students' notes, considering standard textbooks as the reference. The most significant purpose of the latter method is to make use of textbook knowledge structured by humans, thereby increasing the quality of the knowledge base. The following sections elaborate on the details of the methodology and implementation.

\section{Manufacturing Knowledge Graph Construction}

\subsection{KG construction using Wikidata}

Wikidata is a knowledge base maintained collaboratively by the community to represent information in machine readable format. Since no such knowledge base exists for the manufacturing domain, we decided first to extract existing Wikidata knowledge and then merge this with the knowledge contained within manufacturing textbooks.

Wikidata’s knowledge graph has Q and P identifiers where Q represents entities, and P represents relations \cite{hernandez2015reifying}. Currently, Wikidata is limited to a very few relevant relations between entities when it comes to manufacturing domain specific entities. We have taken about 10 unique relations based on all P identifiers attached with relevant Q identifiers identified by us. The relations include `Instance of', `Subclass of', `Use', `Color', `Part of', `Uses', `Has quality', `Has cause', `Has part', `Facet of', `Different from'.

In order to find manufacturing-specific entities in Wikidata, we used the following methods:

\subsubsection{Entities extraction from index words of textbooks}
Index words located at the end of textbooks are a list of all topics and entities provided to assist readers in finding the location of the text. These are important terms that are often overlooked but are a good collection of domain-specific entities. We utilized easily accessible 5 diverse ebooks related to manufacturing \cite{groover2020fundamentals}, digital manufacturing \cite{zhou2012fundamentals}, manufacturing process \cite{el2005advanced} welding technology \cite{kou2003welding} and additive manufacturing \cite{gibson2021additive}, and extracted the index entities mentioned at the end of the book to expand the list of relevant entities. We found about 3500 relevant entities from various books and added those to our vocabulary.

\subsubsection{Keywords from research papers}
We used 500k+ abstracts to create the corpus for manufacturing, as mentioned in FabNER. While extracting the abstracts, we accumulated the keywords mentioned in the abstract, removed duplicates, and normalized many of the words (using Levenshtein distance). There are many words written with some variation in the spelling. E.g., Landau-Ginzburg-Devonshire, Landau-Ginsburg-Devonshire, Landau-Ginsberg-Devonshire, are the same entities with variation in the way it is written in different abstract keywords by various authors. Overall, we found about 4500 relevant entities from a sample of 5000 abstracts.

\subsubsection{Named entity recognition on unstructured text}
We utilized review articles related to manufacturing to find the most frequent and diverse terms since it generally mention most of the past work and technologies developed in the succinct text. Ten full review articles \cite{wong2012review, elmaraghy2012complexity, zhu2013review, frazier2014metal, oztemel2020literature, yan2018review, stuart2010emerging, rajurkar2017review, wang2020intelligent, kaur2019review} for this part were selected, which were processed using a trained neural network model consisting of BERT \cite{devlin2018bert} and GloVe \cite{pennington2014glove} stacked embeddings through Flair framework \cite{akbik2019flair}. Next, we employed BiLSTM and CRF \cite{consoli2019multidomain} architecture to identify 12 category entities in the review articles with F-score of 83\%. Overall, we found about 2000 entities from diverse review articles related to manufacturing. 

\begin{table}[htb]
\caption{Named entity recognition performance for Manufacturing dataset}
\resizebox{\columnwidth}{!}{%
\begin{tabular}{|c c c c|} 
 \hline
 Model & Precision & Recall & F1 \\ [0.5ex] 
 \hline
 BERT+BiLSTM+CRF & 0.8185 & 0.8429 & 0.8306 \\ 
 \hline
\end{tabular}
}
\end{table}

Using text and vocabulary of entities from all the above sources, i.e., index words, research paper keywords, and NER on review articles, we further employed two methods for finding existing Wikidata items. As depicted in Fig. \ref{figure1}, in the first method, we used DBpedia spotlight API to find Wikidata items associated with the unstructured text directly based on a 0.5 confidence value. In the second, we provide manufacturing vocabulary terms as the input to wptools python library to fetch Wikidata items as the output. We find all manufacturing relevant Wikidata items to extract a subgraph from Wikidata and later merge this relatively bigger knowledge graph with textbook knowledge (explained in the next section). Upon availability of some Wikidata items, we further used SPARQLWrapper (uses Wikidata SPARQL endpoint) and relations list (P identifiers) to fetch forward (head from the primary entity) as well as backward (tail from the primary entity) entities associated with the item. We performed the same for two linked steps forward and two linked steps backward to find most of the nodes that are connected with each other.

\begin{figure*}
    \centering
    \includegraphics[width=\textwidth,height=7cm]{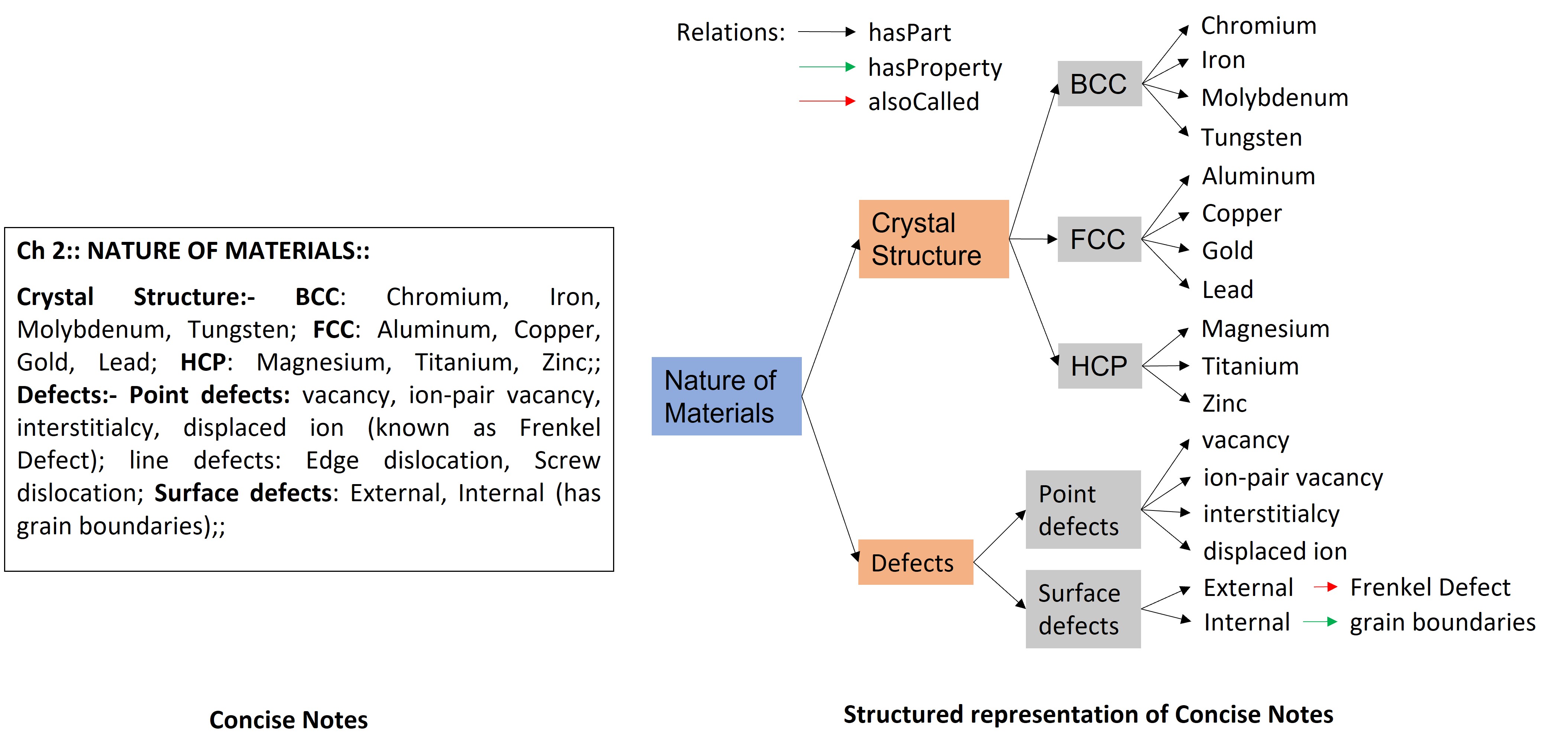}
    \caption{Conversion of concise notes to structured graph}
    \label{figure2}
\end{figure*}

\subsection{KG construction using Exam cheatsheet/notes for Manufacturing}
We propose a novel approach for creating triples utilizing human knowledge. Qualifying exams (or course exams) are part of any doctoral degree program. In some schools, written exams are conducted for a few courses. In some specific courses, cheatsheets/concise notes are allowed for students to bring into the exam to enable the student to remember important points. In most cases, the cheatsheet (or notes) developed for the exam are useless when the exams are over. This also means that the verified knowledge written by a student to remember the essential facts is lost or left unutilized for future references. 

We devised a strategy for making these short, concise notes be useful input for building connected entities within FabKG. We created optional advice on cheatsheet generation for students to follow prior to the exam at our institution so that they could participate to the task of knowledge base enhancement in the Manufacturing area. Students can only write crucial details from various textbook chapters, assuming that the number of pages allowed in the exam is limited. There is a title within each chapter, followed by several subtitles, each of which contains some entities and context, which is potentially a good knowledge source. The guidelines were kept simple so that students would not have to spend much time referring to them. It mentioned the title, subtitle, and content hierarchy and a precise technique for separating them. 

The following guidelines are provided for example purposes only:
a) The chapter name is preserved as the top title, followed by a distinctive symbol, making it easier to distinguish between chapters.
b) Within a chapter, many sub-topics are separated by another unique symbol, such as a double semi-colon ';;'. Two sub-topics are shown in figure 2, for example: (1) Defects and (2) Crystal structure
c) If there is a further subtopic within a subtopic, it is separated by a symbol such as ':' followed by some relevant points. A single semi-colon separates multiple subtopics.
d) Explanations or additional information about any term are retained in brackets as an attribute of a relational entity. For example, displaced ion (Frenkel defect) denotes that a point defect with a displaced ion is also known as a Frenkel defect.

Use of some symbols patterns when creating the notes aided in the design of regex patterns for quickly extracting entities and their obvious relationships. We were able to extract over 1200 distinct entities, 25 unique relations, and 4200 unique triples using this method. Fig. \ref{figure2} depicts the notes in their raw and structured state. The student notes in both unstructured and structured form was verified by human supervision. Indirect crowdsourcing is the crucial aspect that has made this element of the project possible. However, the intention was to use note takers' knowledge. It should be emphasized that even though some previous work has mentioned the use of notes \cite{denny2015using} for developing a knowledge map, on a larger scale and for educational applications, this type of knowledge source has not been studied. This method might be used with little effort for any domain-specific textual material.

Despite the small number of entities/relations discovered, this method allows textbook knowledge to be converted into useable knowledge, which aids in developing a knowledge graph for educational purposes. In general, for automatic extraction of directed relation, it is often difficult to determine which entities are related to each other when more than 2 entities are present in a sentence. This is also because, on multiple occasions, no relation exists between the entities. It becomes a challenge to employ a NER and detect directed relations between entities automatically which we solve by this semi-supervised method. Based on the analysis of the notes, some of the crucial relations found include: `has', `hasProperty', `uses', `usedTo', `usedIn', `causes', `producedBy', `makes', `hasExpression', `hasPart', `addedWith', `hasValue', `includes', `partOf', `alsoCalled', `dueTo', `instanceOf', `isAbbrev', `isAcronym', `hasComparator'.

\subsection{Fusion of structured and unstructured knowledge}
All triples found with the above-mentioned methods were aggregated together to create a knowledge graph of about 65000 triples. Fig\ref{figure1}(b) depicts the merger of Wikidata and textbook knowledge. We created a collection of possible synonyms for various entities to enable us to merge Wikidata entities with textbook entities. We found that out of 1200 textbook entities, about 25\% were present in Wikidata. We also found some links between entities which otherwise were not present in Wikidata due to limited relations.

\section{Knowledge driven QA}

\subsection{Domain specific question answering}

The Knowledge Graph for manufacturing (FabKG) is suitable for answering questions and powering a chatbot to answer questions. The FabKG is a directed graph G = (V, E) where the node v $\in$ V denotes named entities of manufacturing, numeric literal or expression, and the edge e $\in$ E denotes directed relation between the nodes.

Given a natural language question as input, the entities are categorized in their respective classes. Based on the subject and predicate most similar object (highest cosine similarity) to the category in the knowledge base is queried. 

Some of the common domain specific questions could not be answered using general purpose search engines. Examples of questions that could be answered by FabKG are:
a.	Which tool geometry is used for planning?
b.	Which material has more hardness, cermet or alumina?
Note: We have used a hasComparator relation specifying various comparison values in our KG that could answer the ‘more’ and ‘less’ inference question.
c.	What is the composition of Tungsten in cast cobalt?
d.	Which nontraditional manufacturing process is used for coining operations?
e.	What is the length to depth ratio for discontinuous fibers?

\subsection{Expression based question answering}
We have included some manufacturing-specific formulas/expressions in the knowledge graph to enable inference-based calculations.
Since we have captured some formulae linked with entities using ‘hasExpression’ relation, traversing for the formula node in the graph is easy. We have also included a simple rule for calculation-type questions. Here is an example question below:

\begin{quote}
Calculate the strain on the cylinder given the area 1 cm$^2$, 10N force, and Young's modulus for steel 200 GPa.
\end{quote}

\begin{figure}
    
    \includegraphics[width=\linewidth,height=4cm]{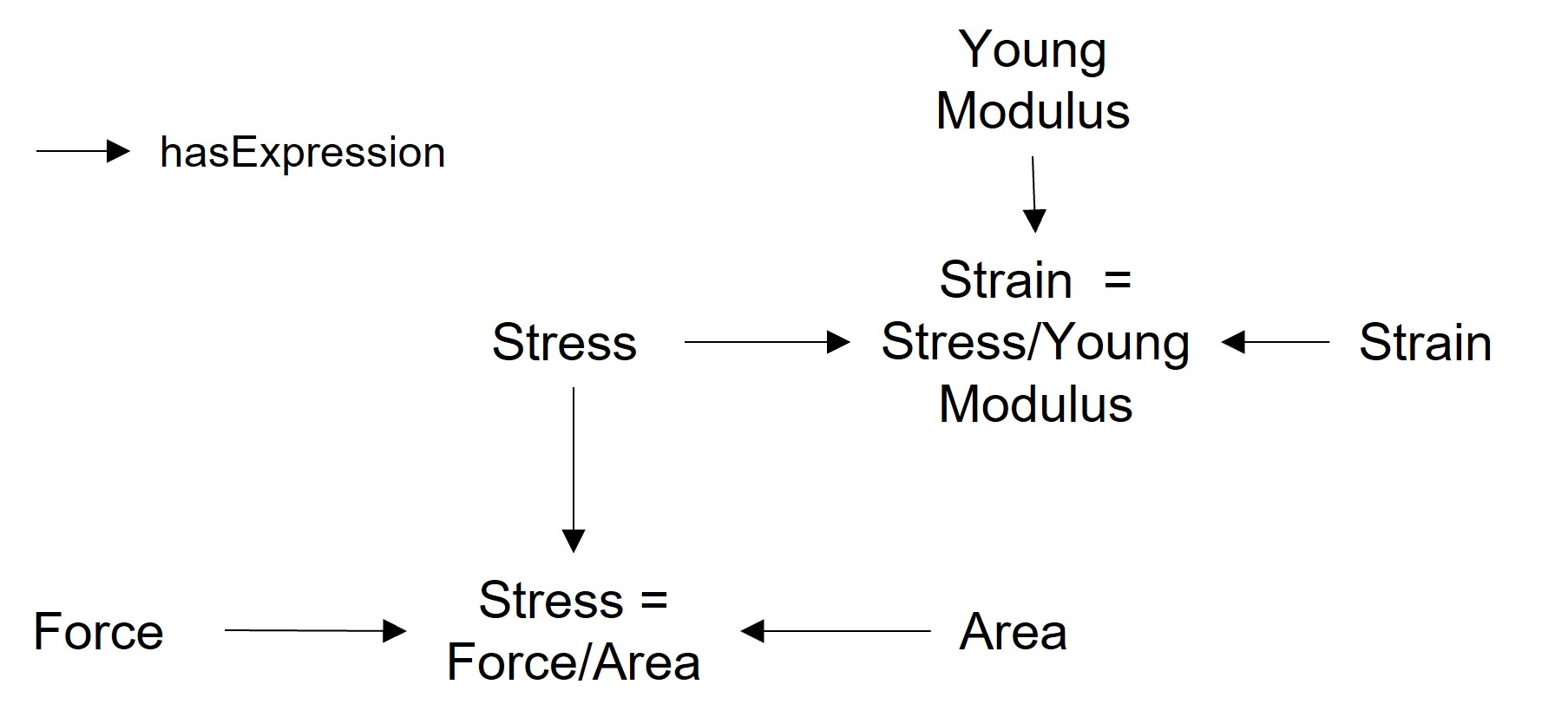}
    \caption{A small subgraph showing the links of entities connected with other expressions for ease of calculation, making the system think the we way just like humans do. }
    \label{figure3}
\end{figure}

Given the question above, we have some `formula entities': area, force, and young modulus of steel. These entities are queried in the KG for any available linked expression. Similar to MathGraph \cite{zhao2019mathgraph}, we utilize SymPy \cite{meurer2017sympy} to convert the queried expression into a mathematical equation with variables, and to perform the calculation, we use some basic rules of precedence to fetch the results. As shown in fig. \ref{figure3}, we can find strain using Young's modulus and stress; however, since stress is not known, we calculate stress as the first step using force and area. This process depicts the way human thinks while answering a question with some inputs and related expressions.

Some other examples of questions forms that are easier than the above-mentioned questions:
a.	Calculate material removal rate given feed rate, cutting speed, and depth of cut. HINT: We can calculate the Material removal rate using (feed rate)*(cutting speed)*(depth of cut).
b.	Calculate measuring length of roughness given cutoff length of 0.8. HINT: measuring length of roughness = 0.5 * cutoff length.

\section{Conclusion and future work}

We have developed FabKG – a knowledge graph for product design and manufacturing, which utilizes two critical sources of knowledge, (1) Wikidata and (2) Human constructed notes, that combine structured/unstructured knowledge towards answering question-related to product development and manufacturing. Using this KG, students, product developers, and knowledge seekers can get good insights into various concepts and fundamentals about various topics in this domain. Using all the methods described above, we have found 65000+ triples in 12 entity categories. In the future, we plan to use the heterogeneous knowledge graph for directed relation prediction in the bigger corpus, performing graph embedding and link prediction. Moreover, lecture presentations with succinct text could also be utilized for finding entities and relations. Generally, the title/topic of the presentation symbolizes the subject, with some entities either written directly or placed after another subtopic.
Furthermore, ‘property/attribute’ of relation through the specific value of entities such as the strength of materials, carbon content, Brinell hardness, Etc., currently available in tabular form in books and other resources, can be added to the KG. The same could be represented using a hypergraph by combining multimodal data. Therefore, the new graph structure would have not only an ‘entity-relation-entity’ type graph but also an ‘entity-attribute-value’ graph. Finally, this knowledge graph could help link to global knowledge by contributing to existing Wikidata knowledge with the help of Wikimapper.

\bibliography{anthology,custom}
\bibliographystyle{acl_natbib}

\end{document}